\documentclass[review]{elsarticle}

\usepackage[export]{adjustbox}
\usepackage{caption}
\usepackage{subcaption}

\usepackage{placeins}

\usepackage[nolist]{acronym}

\usepackage{longtable}
\usepackage{booktabs}
\usepackage{multirow}
\usepackage{float}

\usepackage{gensymb}

\usepackage[hidelinks]{hyperref}

\usepackage{lineno}
\modulolinenumbers[5]

\journal{Journal of \LaTeX\ Templates}

\newcommand{\specialcell}[2][l]{%
  \begin{tabular}[#1]{@{}l@{}}#2\end{tabular}}

\newcommand{\mytilde}{\raise.17ex\hbox{$\scriptstyle\mathtt{\sim}$}}

\biboptions{authoryear}

\begin{document}

\begin{frontmatter}

	\title{Performance evaluation and hyperparameter tuning of statistical and machine-learning models using spatial data}

	\author[FSU]{Patrick Schratz}
	\cortext[mycorrespondingauthor]{Corresponding author}
	\ead{patrick.schratz@uni-jena.de}

	\author[FSU]{Jannes Muenchow}
	\author[NEIKER]{Eugenia Iturritxa}
	\author[TUDO]{Jakob Richter}
	\author[FSU]{Alexander Brenning}

	\address[FSU]{Department of Geography, GIScience group, Grietgasse 6, 07743, Jena, Germany}
	\address[NEIKER]{NEIKER, Granja Modelo –Arkaute, Apdo. 46, 01080 Vitoria-Gasteiz, Arab, Spain}
	\address[TUDO]{Department of Statistics, TU Dortmund University, Germany}

	\begin{abstract}
		Machine-learning algorithms have gained popularity in recent years in the field of ecological modeling due to their promising results in predictive performance of classification problems.
		While the application of such algorithms has been highly simplified in the last years due to their well-documented integration in commonly used statistical programming languages such as R, there are several practical challenges in the field of ecological modeling related to unbiased performance estimation, optimization of algorithms using hyperparameter tuning and spatial autocorrelation.
		We address these issues in the comparison of several widely used machine-learning algorithms such as Boosted Regression Trees (BRT), k-Nearest Neighbor (WKNN), Random Forest (RF) and Support Vector Machine (SVM) to traditional parametric algorithms such as logistic regression (GLM) and semi-parametric ones like Generalized Additive Models (GAM).
		Different nested cross-validation methods including hyperparameter tuning methods are used to evaluate model performances with the aim to receive bias-reduced performance estimates.
		As a case study the spatial distribution of forest disease (\textit{Diplodia sapinea}) in the Basque Country in Spain is investigated using common environmental variables such as temperature, precipitation, soil or lithology as predictors.

		Results show that GAM and \ac{RF} (mean AUROC estimates 0.708 and 0.699) outperform all other methods in predictive accuracy.
		The effect of hyperparameter tuning saturates at around 50 iterations for this data set.
		The AUROC differences between the bias-reduced (spatial cross-validation) and overoptimistic (non-spatial cross-validation) performance estimates of the GAM and RF are 0.167 (24\%) and 0.213 (30\%), respectively.
		It is recommended to also use spatial partitioning for cross-validation hyperparameter tuning of spatial data.
		The models developed in this study enhance the detection of \textit{Diplodia sapinea} in the Basque Country compared to previous studies.
	\end{abstract}

	\begin{keyword}
		spatial modeling \sep machine learning \sep model selection \sep hyperparameter tuning \sep spatial cross-validation
	\end{keyword}

\end{frontmatter}


\begin{acronym}[AUROC]

	\acro{ANN}{Artificial Neural Network}
	\acro{AUROC}{Area Under the Receiver Operating Characteristics Curve}
	\acro{BRT}{Boosted Regression Trees}
	\acro{CART}{Classification and Regression Trees}
	\acro{CV}{cross-validation}
	\acro{ENM}{Environmental Niche Modeling}
	\acro{FPR}{False Positive Rate}
	\acro{GAM}{Generalized Additive Model}
	\acro{GBM}{Gradient Boosting Machine}
	\acro{GLM}{Generalized Linear Model}
	\acro{IQR}{Interquartile Range}
	\acro{WKNN}{Weighted $k$-nearest neighbor}
	\acro{MARS}{Multivariate Adaptive Regression Splines}
	\acro{MEM}{Maximum Entropy Model}
	\acro{LOWESS}{Locally Weighted Scatter Plot Smoothing}
	\acro{PISR}{Potential Incoming Solar Radiation}
	\acro{RBF}{Radial Basis Function}
	\acro{RF}{Random Forest}
	\acro{SDM}{Species Distribution Modeling}
	\acro{SVM}{Support Vector Machines}
	\acro{TPR}{True Positive Rate}
\end{acronym}

\section{Introduction}
\label{sec:intro}
Statistical learning has become an important tool in the process of knowledge discovery from big data in fields as diverse as finance or geomarketing \citep{Heaton2016, Schernthanner2017}, medicine \citep{Leung2016}, public administration \citep{Maenner2016} and the sciences \citep{Garofalo2016}.
We can classify statistical learning broadly into supervised and unsupervised techniques (e.g., ordination, clustering) \citep{James2013}.
Though both fields are important in the spatial modeling field, we will focus in this paper on supervised predictive modeling and the comparison of (semi-)parametric models and machine learning techniques.
Spatial predictions are of great importance in a wide variety of fields including geomorphology \citep{brenning_landslide_2015}, remote sensing \citep{Stelmaszczuk2017}, hydrology \citep{Naghibi2016}, epidemiology \citep{Alder2017}, climatology \citep{Voyant2017}, the soil sciences \citep{Hengl2017} and of course ecology.
Ecological applications range from species distribution models \citep{Halvorsen2016, Quillfeldt2017, Wieland2017}, predicting floristic \citep{muenchow_coupling_2013} and faunal composition to disentangling the relationships between species and their environment \citep{muenchow_soil_2013}.
Additional applications include biomass estimation \citep{Fassnacht2014} and disease mapping as for example caused by fungal infections \citep{Iturritxa2014}.
The latter marks the research area of this work.

Fungal species such as \textit{Diplodia sapinea} inflict severe damage upon Monterrey pine trees (\textit{Pinus radiata}) which trees are subjected to environmental stress \citep{Wingfield2008}.
Infected forest stands cause economic as well as ecological damages worldwide \citep{Ganley2009}.
In Spain, where timber production is regionally an important economic factor, about 25\% of the timber production stems from Monterrey pine (\textit{Pinus radiata}) plantations in northern Spain, and here mostly from the Basque Country \citep{Iturritxa2014}.
Consequently, the early detection and subsequent containment of fungal diseases is of great importance.
Statistical and machine-learning models play an important role in this process.

Supervised techniques can be broadly divided into parametric and non-parametric models.
Parametric models can be written as mathematical equations involving model coefficients.
This enables ecologists to interpret interactions between the response and its predictors and to improve the general understanding of the modeled relationship.
Model interpretability should certainly be an important criterion for choosing models when the analysis of relationships between a response variable such as species richness or species presence/absence and the corresponding environment is of interest \citep{Goetz2015}.
While the most commonly used statistical models such as generalized linear models (GLMs) are parametric, especially machine learning techniques offer a non-parametric approach to spatial modeling in ecology.
These have gained popularity due to their ability to handle high-dimensional and highly correlated data and the lack of explicit model assumptions.
Some model comparison studies in the spatial modeling field suggest that machine learning models might be the better choice when the primary aim is accurate prediction \citep{Hong2015, Smolinski2016, Youssef2015}.
However, other studies found no major performance difference to parametric models \citep{Bui2015, Goetz2015}.

The estimation of predictive performances and the tuning of model hyperparameters (where present) are two intertwined critical issues in ecological modeling and model comparisons, both of which are addressed in this study.
Cross-validation and bootstrapping are two widely used performance estimation techniques \citep{Brenning2005, Kohavi1995}.
However, in the presence of spatial autocorrelation, estimates obtained using regular (non-spatial) random resampling may be biased and overoptimistic, which has led to the adoption of spatial resampling in cross-validation and bootstrapping for bias reduction.
Currently, different names are used in science for the same idea: \cite{Brenning2005} named it "spatial cross-validation", \cite{Meyer2018} "Leave-location-out cross-validation" and \cite{Roberts2017} labels it "Block cross-validation".
Although the importance of bias-reduced spatial resampling methods for performance estimation has been emphasized repeatedly in recent years \citep{Geiss2017, Meyer2018, Wenger2012}, such techniques have not been adopted in all cases  \citep{Bui2015, Pourghasemi2018, Smolinski2016, Wollan2008, Youssef2015}.
Since default hyperparameter settings, which are used by some authors \citep{Goetz2015, Russ2010a, Russ2010b, Vorpahl2012}, can in no way guarantee an optimal performance of machine-learning techniques, additional attention should be directed to this potentially critical step.
Again, performance estimation techniques such as cross-validation are used in this step, and the adequacy of non-spatial techniques for spatial data sets can be questioned.
This work aims to be an exemplary model comparison study for spatial data using spatial cross-validation including spatial hyperparameter tuning to receive bias-reduced performance estimates.
This approach is compared with cross-validation approaches that use other resampling strategies (i.e. random resampling) or conduct no hyperparameter tuning.

We provide the complete code (including a packrat file) in the supplementary material to make this work fully reproducible and to encourage a wider adoption of the proposed methodology.
In our exemplary analysis we used a selection of six models (statistical and machine-learning) that are commonly used in the spatial modeling field: \ac{BRT}, \ac{GAM}, \ac{GLM}, \ac{WKNN}, \ac{RF} and \ac{SVM}.

\section{Data and study area}

\begin{figure} [t!]
	\begin{center}
		\makebox[\textwidth]{\includegraphics[width=\textwidth] {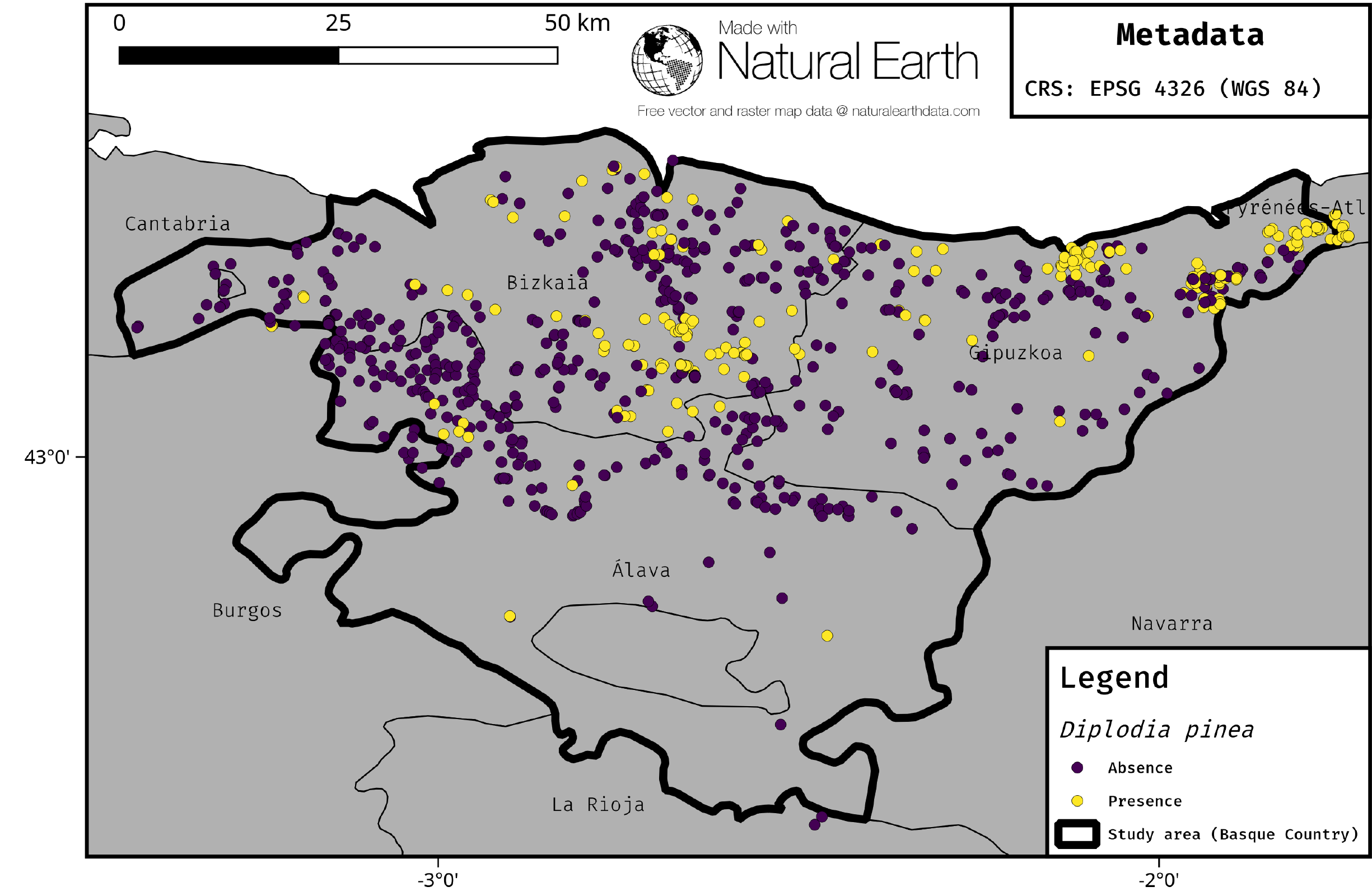}}
		\caption[Study area]{Spatial distribution of tree observations within the Basque Country, northern Spain, showing infection state by \textit{Diplodia sapinea}.}
		\label{fig: study_area}
	\end{center}
\end{figure}

\subsection{Data}

This study uses the data set from \cite{Iturritxa2014} to illustrate procedures and challenges that are common to many geospatial analyses problems:
An uneven distribution of the binary response variable, influence of spatial autocorrelation and predictor variables derived from various sources (other modeling results, remote sensing data, surveyed information).
It is representative for many other ecological data sets in terms of sample size (926) and the number (11) and types of predictors (numeric as well as nominal).
The following (environmental) variables were used as predictors: mean temperature (March - September), mean total precipitation (July - September), \ac{PISR}, elevation, slope (degrees), potential hail damage at trees, tree age, pH value of soil, soil type, lithology type, and the year when the tree was surveyed.
Tree infection caused by fungal pathogens (here \textit{Diplodia sapinea}) represents the response variable.
The ratio of infected and non-infected trees in the sample is roughly 1:3 (223, 703).
Compared to the original data set from \cite{Iturritxa2014}, we added soil type (aggregated from 12 to 7 classes in accordance with the world reference base \citep{wrb2014}) \citep{Hengl2017}, lithology type (condensed from 17 to 5 classes) \citep{lithology} and pH value of the soil \citep{ph} to the already available predictors.

\cite{Iturritxa2014} showed that hail damage explained best pathogen infections in trees in the Basque Country. 
In this study hail damage was a binary predictor available as in-situ observations.
To make it available as a predictor for the Basque country, we spatially predicted the hail damage potential as a function of climatic variables using a \ac{GAM} \citep{Schratz2016}.

Predictor \textit{soil} was predicted by \cite{Hengl2017} using ca. 150.000 soil profiles at a spatial resolution of 250 m.
Predictor \textit{age} was imputed and trimmed to a value of 40 to reduce the influence of outliers.
Predictor \textit{pH} was mapped by \cite{ph} using a regression-kriging approach based on 12,333 soil pH measurements from 11 different sources.
Spatial predictions utilized 54 auxiliary variables in the form of raster maps at a 1\,km $\times$ 1\,km resolution and were aggregated to a spatial resolution of 5\,km $\times$ 5\,km.
Information about lithology types were extracted from a classification provided by GeoEuskadi that is based on the year 1999 \citep{lithology}.
Rock type condensing was done using the respective top level class for magmatic types and sub-classes for sedimentary rocks \citep{Grotzinger2016} (\autoref{table:descriptive_summary_non_numeric}).

We removed three observations due to missing information in some variables leaving a total of 926 observations (\autoref{table:descriptive_summary_numeric}).
The methodology we present in this work, i.e. a binary classification problem, can be easily adapted to multiclass problems as well as to quantitative response variables.

\subsection{Study area}

The Basque country in northern Spain represents our study area (\autoref{fig: study_area}).
It has a spatial extent of 7355 km\textsuperscript{2}.
Precipitation decreases towards the south while the duration of summer drought increases.
Between 1961 and 1990, mean annual precipitation ranged from 600 to 2000 mm\, with annual mean temperatures between 8 and 16\degree C \citep{Ganuza2003}.
The wooded area covers approximately 54\% of the territory (396.962 hectars), which is one of the highest ratios in the EU.
Radiata pine is the most abundant species occupying 33.27\% of the total area \citep{Mugica2016}.

\section{Methods}

In this study we provide an exemplary analysis combining both tuning of hyperparameters using nested \ac{CV} and the use of spatial \ac{CV} to assess bias-reduced model performances.
We compared predictive performances using four setups: non-spatial \ac{CV} for performance estimation combined with non-spatial hyperparameter tuning (\emph{non-spatial/non-spatial}), spatial \ac{CV} estimation with spatial hyperparameter tuning (\emph{spatial/spatial}), spatial \ac{CV} estimation with non-spatial hyperparameter tuning (\emph{spatial/non-spatial}), and spatial \ac{CV} estimation without hyperparameter tuning (\emph{spatial/no tuning}).
We used a selection of commonly used machine learning algorithms (\ac{RF}, \ac{SVM}, \ac{WKNN}, \ac{BRT}) and the statistical methods \ac{GLM} and \ac{GAM}.

\subsection{Cross-validation estimation of predictive performance}
\label{sec:nested_cv}

\begin{figure} [t!]
	\begin{center}
		\makebox[\textwidth]{\includegraphics[width=\textwidth] {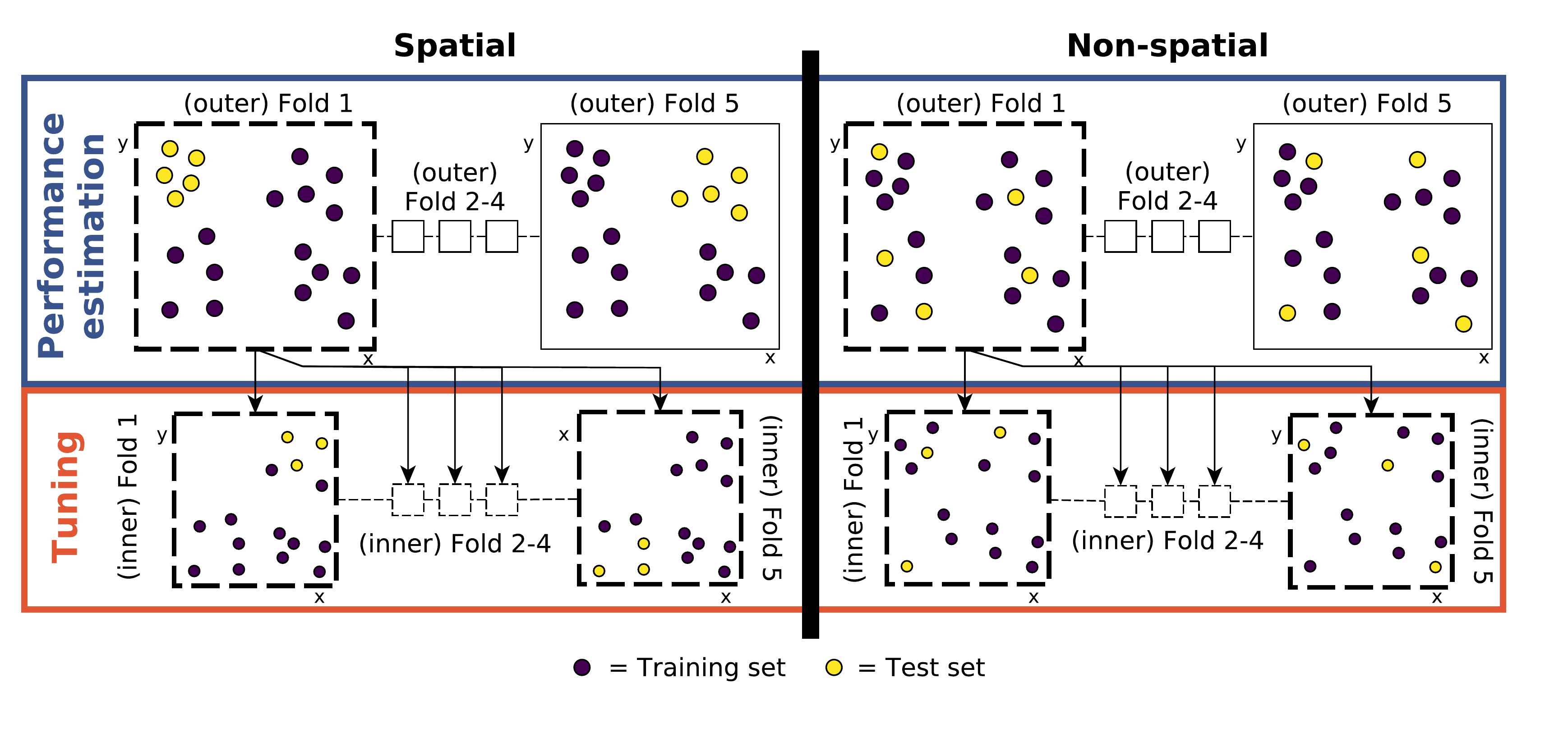}}
		\caption[]{Theoretical concept of spatial and non-spatial nested cross-validation using five folds for hyperparameter tuning and performance estimation.
			Yellow/purple dots represent the training and test set for performance estimation, respectively.
			The tuning sample is based on the respective performance estimation fold sample and consists again of training (orange) and test set (blue).
			Although the tuning folds of only one fold are shown here, the tuning is performed for every fold of the performance estimation level.}
		\label{fig:nested_cv}
	\end{center}
\end{figure}

Cross-validation is a resampling-based technique for the estimation of a model's predictive performance \citep{James2013}.
The basic idea behind \ac{CV} is to split an existing data set into training and test sets using a user-defined number of partitions (\autoref{fig:nested_cv}).
First, the data set is divided into $k$ partitions or folds.
The training set consists of $k - 1$ partitions and the test set of the remaining partition.
The model is trained on the training set and evaluated on the test partition.
A repetition consists of $k$ iterations for which every time a model is trained on the training set and evaluated on the test set.
Each partition serves as a test set once.

In ecology, observations are often spatially dependent \citep{Dormann2007, Legendre1989}.
Subsequently, they are affected by underlying spatial autocorrelation by a varying magnitude \citep{Brenning2005, Telford2005}.
Model performance estimates should be expected to be overoptimistic due to the similarity of training and test data in a non-spatial partitioning setup when using any kind of cross-validation for tuning or validation \citep{sperrorest}.
Therefore, cross-validation approaches that adapt to this problem should be used in any kind of performance evaluation when spatial data is involved \citep{sperrorest, Meyer2018, Telford2009}.
In this work we use the spatial cross-validation approach after \cite{sperrorest} which uses $k$-means clustering to reduce the influence of spatial autocorrelation.
In contrast to non-spatial CV, spatial CV reduces the influence of spatial autocorrelation by partitioning the data into spatially disjoint subsets (\autoref{fig:nested_cv}).

Five-fold partitioning repeated 100 times was chosen for performance estimation (\autoref{fig:nested_cv}).
For the hyperparameter tuning, again five folds were used to split the training set of each fold.
Hyperparameter tuning only applied to the machine learning algorithms.
A random search with a varying number of iterations (0, 10, 50, 100, 200) was applied to each fold of the tuning level.
Model performances of every hyperparameter setting were computed at the tuning level and averaged across folds.
The hyperparameter setting with the highest mean \ac{AUROC} result across all tuning folds was used to train a model on the training set of the respective performance estimation level.
This model was then evaluated on the test set of the respective fold (performance estimation level).
The procedure was repeated 500 times (100 repetitions with five folds each and varying random search iterations) to reduce the variance introduced by partitioning.

The \ac{AUROC} was selected as a goodness of fit measure due to the binary response variable.
The present methodology can also be applied with other measures than AUROC which are suited for binary classification.
This measure combines both \ac{TPR} and \ac{FPR} of the classification and is also independent of a specific decision threshold \citep{Candy2013}.
A resulting \ac{AUROC} value of close to 0.5 indicates no separation power of the model while a value of 1.0 would mean that all cases were correctly classified.

Hyperparameter tuning was performed for \ac{RF}, \ac{SVM}, \ac{BRT} and \ac{WKNN}.
For \ac{GLM}, no tuning is needed because the model has no hyperparameters and assumes a logit relationship between response and predictors.
For \ac{GAM}, see \autoref{subsec:gam}.

\subsection{Tuning of hyperparameters}
\label{subsec:methods_tuning}

\begin{table}[b!]
\centering
\caption[t]{Hyperparameter limits and types for each model.
	Notations of hyperparameters from the respective R packages were used.}
\begingroup\footnotesize
\begin{tabular}{llllrr}
	\\
	Algorithm (package)            & Hyperparameter             & Type    & Value & Start     & End      \\
	\hline
	\multirow{3}{*}{BRT (gbm)}     & \texttt{n.tree}            & integer & -     & 100       & 10000    \\
	                               & \texttt{shrinkage}         & numeric & -     & 0         & 1.5      \\
	                               & \texttt{interaction.depth} & integer & -     & 1         & 40       \\
	\midrule
	\multirow{2}{*}{RF (ranger)}   & \texttt{mtry}              & integer & -     & 1         & 11       \\
	                               & \texttt{num.trees}         & integer & -     & 10        & 10000    \\
	\midrule
	\multirow{2}{*}{SVM (kernlab)} & \texttt{C}                 & numeric & -     & $2^{-12}$ & $2^{15}$ \\
	                               & \texttt{$\sigma$}          & numeric & -     & $2^{-15}$ & $2^{6}$  \\
	\midrule
	\multirow{3}{*}{WKNN (kknn)}   & \texttt{k}                 & integer & -     & 10        & 400      \\
	                               & \texttt{distance}          & integer & -     & 1         & 100      \\
	                               & \texttt{kernel}            & nominal & *     &           &          \\
	\bottomrule
	\multicolumn{6}{l}{* triangular, Epanechnikov, biweight, triweight, cos, inv, Gaussian, optimal}     \\
\end{tabular}
\endgroup
\label{tab:hyperparameter_limits}
\end{table}

Determining the optimal (hyperparameter) settings for each model is crucial for the bias-reduced assessment of a model's predictive power.
While (semi-)parametric algorithms cannot be tuned in the same way as machine-learning algorithms (although some perform an internal optimization, e.g. the implementation of the GAM in the \textit{mgcv} package from \cite{mgcv}), hyperparameters of machine-learning algorithms need to be tuned to achieve optimal performances \citep{Bergstra2012, Duarte2017, Hutter2011}.
Note that for parametric models the term "parameter" is often used to refer to the regression coefficients of each predictor in the fitted model.
For machine-learning algorithms, the terms "parameter" and "hyperparameter" both refer to "hyperparameter" as there are no regression coefficients for these models.
In addition, the term "parameter" is often used in programming to refer to an argument of a function.
These different usages often lead to confusion and hence both terms should be used with caution.
Hyperparameters are determined by finding the optimal value for a model across multiple unknown data sets by using a optimization procedure such as \ac{CV} or Bayesian optimization while parameters of parametric models are estimated when fitting them to the data \citep{Kuhn2013}.

We used a random search with a varying number of iterations (10, 50, 100, 200, 300, 400) for all machine learning models in this study to analyze the difference of varying tuning iterations.
A random search has desirable properties in high dimensional and no disadvantages in low dimensional situations compared to a grid search \citep{Bergstra2012}.
This is due to the fact that often high dimensional situations have a "low effective dimension", i.e. only a subset of the hyperparameters is actually relevant.
Another practical advantage is that one does not have to set the step size for the grid but only the parameter limits.
We did not perform stepwise variable selection or similar for the parametric models (\ac{GLM}, \ac{GAM}) as we required all models to have the same predictor set.
An exploratory analysis was done on using different starting basis dimensions for the optimal smoothing estimation of each predictor of the GAM.
The \textit{mgcv} package does an internal optimization of the smoothing degree value using the supplied basis dimensions as the starting point.
The reported \ac{GAM} model was initiated with $k = 10$ as the basis dimension which ensured full flexibility of the smoothing terms for each predictor.
Please note that although we attributed the GAM to settings \textit{non-spatial/no tuning} and \textit{spatial/no tuning} as we did not perform a tuning ourselves, the GAM actually does a non-spatial optimization of the smoothing degrees for each predictor.
We are aware that this attribution is somewhat contrary to the attribution of all other algorithms in this study.
Strictly, we would also need to implement a spatial optimization procedure of the smoothing degrees for the GAM to follow our philosophy of spatial hyperparameter tuning in this work.
However, such an implementation exceeds the scope of this work.
Belonging to the parametric algorithm group, we decided to attribute the GAM to the "no tuning" class and leave all the tuning settings to the machine-learning models.

All models were fitted using their respective default hyperparameter settings, i.e. no tuning was performed.
For SVM we used $\sigma = 1$ and $C = 1$ to suppress the automatic tuning of the \emph{kernlab} package.
The ranges of the tuning spaces were set by iteratively checking the tuning results and adjusting the search space to make sure that the resulting optimal hyperparameter settings of each fold are not possibly limited by the defined search space.
However, in practice this is sometimes impossible (see the problems we faced for \ac{WKNN} and \ac{BRT} in \autoref{subsec:mod_chars}) because models start to fail if hyperparameter values outside of computationally valid ranges are tested.

Most packages offering \ac{CV} solutions in R offer only random partitioning methods, assuming independence of the observations.
Package \textit{mlr}, which was used as the modeling framework in this work, was missing spatial partitioning functions but provides a unified framework for modeling and simplifies hyperparameter tuning.
With this study we implemented the spatial partitioning methods of package \textit{sperrorest} into \textit{mlr}.

\begin{figure} [t!]
	\begin{center}
		\makebox[\textwidth]{\includegraphics[height = 0.82\textheight] {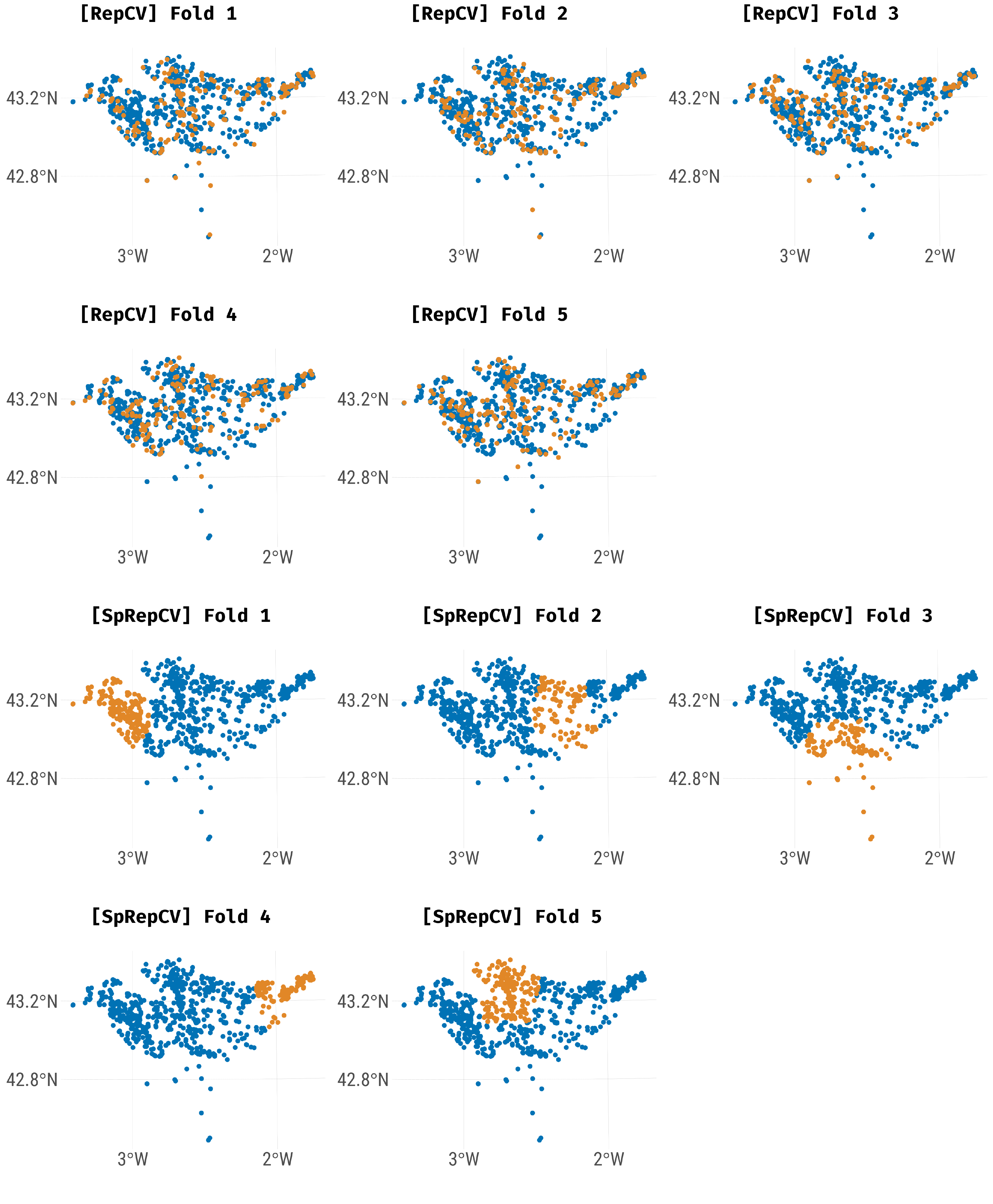}}
		\caption[]{Comparison of spatial and non-spatial partitioning of the first five folds in spatial and non-spatial cross-validation performance estimation.
			Yellow/purple dots represent the training and test set, respectively.}
		\label{fig:cv_settings_comparison}
	\end{center}
\end{figure}

\subsection{Cross-Validation Setups}

To underline the crucial need for spatial \ac{CV} when assessing a model's performance, and to identify overoptimistic outcomes when neglecting to do so, we used following CV setups:
Nested non-spatial \ac{CV} which uses random partitioning and non-spatial hyperparameter tuning (\emph{non-spatial/non-spatial}),
nested spatial \ac{CV} which uses k-means clustering for partitioning \citep{Brenning2005} and results in a spatial grouping of the observations and performs non-spatial hyperparameter tuning (\emph{spatial/non-spatial}) ,
nested spatial \ac{CV} including spatial hyperparameter tuning (\emph{spatial/spatial}) and
spatial \ac{CV} without hyperparameter tuning (\emph{spatial/no tuning}).
Setup (\emph{non-spatial/non-spatial}) was used to show the overoptimistic results when using non-spatial \ac{CV} with spatial data and setups \emph{spatial/non-spatial}, \emph{spatial/spatial} to reveal the differences between spatial and non-spatial hyperparameter tuning.
Setup (\emph{spatial/spatial}) should be used when conducting spatial modeling with machine learning algorithms that require hyperparameter tuning.

\subsection{Model characteristics and hyperparameters}
\label{subsec:mod_chars}

An exemplary selection of widely-used statistical and machine-learning techniques was compared in this study.
While the following sections describe the used models and their settings, a justification of the choice of specific implementations in the statistical software R is included in \autoref{app: A}.
We used the open-source statistical programming language R \citep{R_core} for all analyses and the packages \textit{gbm} \citep{gbm} (\ac{BRT}), \textit{mgcv} \citep{mgcv} (\ac{GAM}), \textit{kernlab} \citep{kernlab} (\ac{SVM}), \textit{kknn} \citep{kknn} (\ac{WKNN}), and \textit{ranger} \citep{ranger} (\ac{RF}).
We have integrated the spatial partitioning functions of the \textit{sperrorest} package into the \textit{mlr} package as part of this work.
\textit{mlr} provides a standardized interface for a wide variety of statistical and machine-learning models in R simplifying essential modeling tasks such as hyperparameter tuning, model performance evaluation and parallelization.

\subsubsection{Random Forest}

Classification trees are a non-linear technique that uses binary decision rules to predict a class based on the given predictors \citep{Gordon1984}.
\ac{RF} aggregates many classification trees by counting the votes of all individual trees.
The class with the most votes wins and will be the predicted class.
Fitting a high number of trees is then referred to as fitting a 'forest' in a metaphorical way.
Using many trees stabilizes the model \citep{Breiman2001}.
However, \ac{RF} saturates at a specific number of trees, meaning that adding more trees will not increase its performance anymore but only increases computing time.
Randomness is introduced in two ways:
First a bootstrap sample ob observations is drawn for each tree.
Second, for each node only a random subset of \texttt{$m_{try}$}) variables is considered for generating the decision rule \citep{Breiman2001}.

\subsubsection{Support Vector Machines}

\ac{SVM}s transform the data in a high-dimensional feature space by performing non-linear transformations of the predictor variables \citep{Vapnik1998}.
In this high-dimensional setting, classes are linearly separated using decision hyperplanes.
The tuning of \ac{SVM}s is important and not trivial due to the sensitivity of the hyperparameters across a wide search space \citep{Duan2003}.

We decided to use the \ac{RBF} kernel (also known as Gaussian kernel) which is the default in most implementations and most commonly used in the literature \citep{e1071, Guo2005, Pradhan2013}.
For this kernel, the regularization parameter \textit{C} and bandwith $\sigma$, which control the degree of non-linearity, are the hyperparameters which have to be optimized.
An exploratory analysis of the Laplace and Bessel kernels was done, which confirmed the expected insensitivity to the choice of the kernel.
All these kernels (including the \ac{RBF} kernel) are classified as "general purpose kernels" \citep{kernlab}.

\subsubsection{Boosted Regression Trees}

\ac{BRT} are different from \ac{RF} in that trees are fitted on top of previous trees instead of being fitted parallel to each other without a relation to adjacent trees.
In this iterative process, each tree learns from the previous fitted trees by a magnitude specified by the \textit{shrinkage} parameter \citep{Elith2008}.
This process is also called 'stage-wise fitting' (not step-wise) because the previous fitted trees remain unchanged while additional trees are added.
\ac{BRT} have a tendency towards overfitting the more trees are added.
Therefore, a combination of a small learning rate with a high number of trees is preferable.
\ac{BRT} acts similar as a \ac{GLM} as it can be applied to several response types (binomial, Poisson, Gaussian, etc.) using a respective link function.
Also, the final model can be seen as a large regression model with every tree being a single term \citep{Elith2008}.
Hyperparameter tuning was performed on the learning rate \textit{shrinkage}, the number of trees \textit{n.tree} and the interaction depth between the variables \textit{interaction.depth}.

\subsubsection{Weighted $k$-Nearest Neighbor}

\ac{WKNN} identifies the K-nearest neighbors within the training set for a new observation to predict the target class based on the majority class among the neighbors.
The first formulation of the algorithm goes back to \cite{Fix1951}.
Besides the standard hyperparameter \emph{number of neighbors} (\textit{$n_{neighbors}$}), the implementation by \cite{kknn} also provides hyperparameter (\textit{distance}) that allows to set the Minkowski distance and a choice between different kernels (up to 12, see \autoref{tab:hyperparameter_limits}).
Hyperparameter \textit{distance} helps finding the $k$-nearest training set vectors which are used for classification together with the maximum of the summed kernel densities provided by hyperparameter \textit{kernel} \citep{kknn}.
Training observations that are closer to the predicted observation get a higher weight in the decision process, when a kernel other then \emph{rectangular} is chosen.
The original idea of the \ac{WKNN} algorithm goes back to \cite{Dudani1976}.

Including weighting and kernel functions may increase predictive accuracy but can also lead to overfitting of the training data.

\subsubsection{Generalized Linear Model and Generalized Additive Models}

\label{subsec:gam}

\ac{GLM}s extend linear models by allowing also non-Gaussian distributions, e.g., binomial, Poisson or negative binomial distributions, for the response variable.
The option to apply a custom link function between the response and the predictors already allows for some degree of non-linearity.
\ac{GAM}s are an extension of \ac{GLM}s allowing the response-predictor relationship to become fully non-linear.
For more details please refer to \cite{Zuur2009, mgcv, James2013}.

\section{Results}

\subsection{Tuning}

\begin{figure} [t!]
	\begin{center}
		\makebox[\textwidth]{\includegraphics[width=0.7\textwidth] {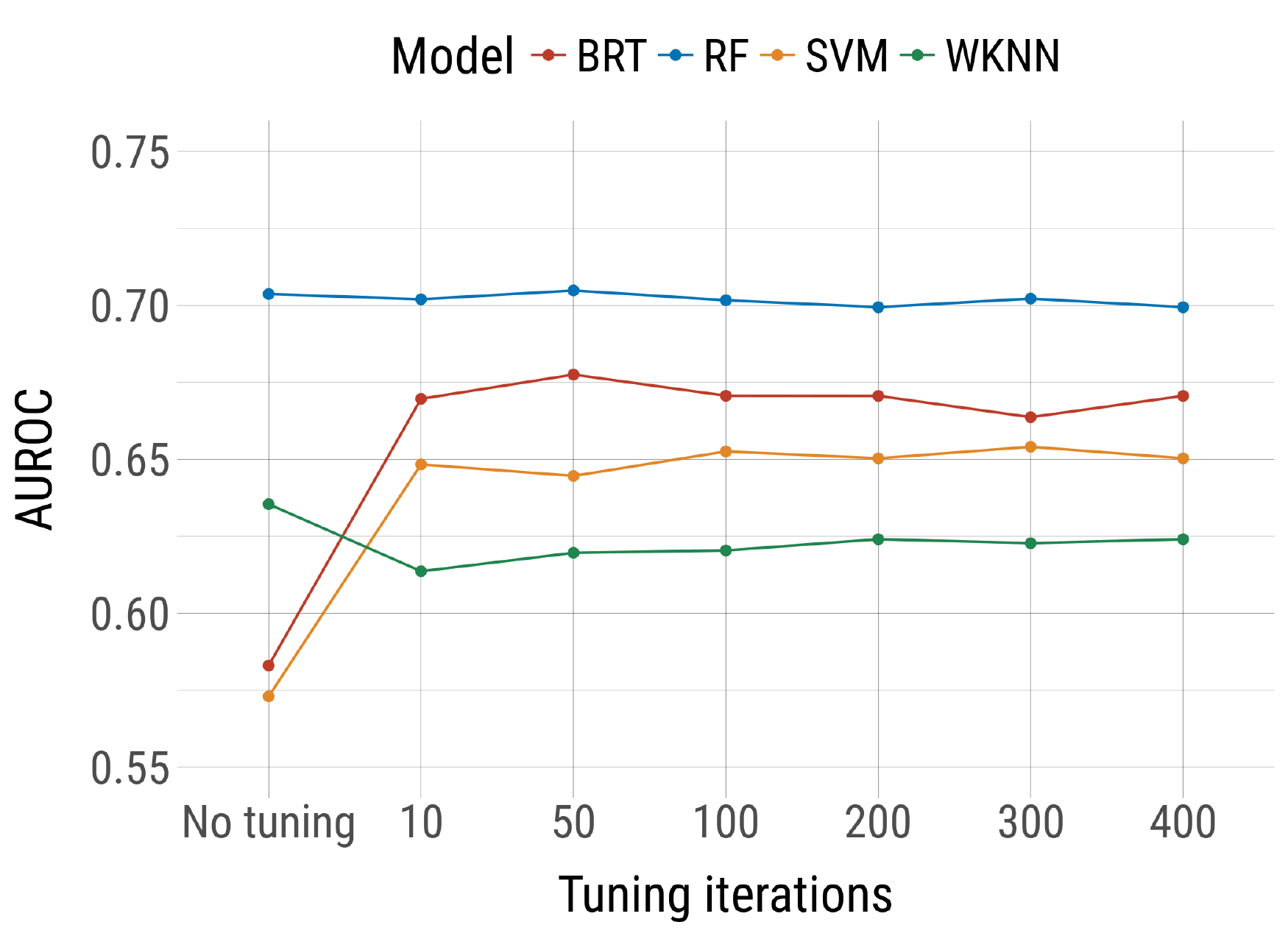}}
		\caption[]{Hyperparameter tuning results of the \emph{spatial/spatial} CV setting for \ac{BRT}, \ac{WKNN}, \ac{RF} and \ac{SVM}: Number of tuning iterations (1 iteration = 1 random hyperparameter setting) vs. predictive performance (AUROC).}
		\label{fig:iterations_vs_auroc}
	\end{center}
\end{figure}


While ten (or more) hyperparameter tuning iterations substantially improved the performance of \ac{BRT} and \ac{SVM} classifiers compared to default hyperparameter values, \ac{WKNN} and \ac{RF} hyperparameter tuning did not result in relevant changes in AUROC (\autoref{fig:iterations_vs_auroc}).
Fifty tuning iterations and more further improved accuracies only slightly (\ac{WKNN}) or not at all (\ac{SVM}, \ac{BRT}).
\ac{SVM} showed the highest tuning effect of all models with an increase of \mytilde 0.08 AUROC (\autoref{fig:iterations_vs_auroc}).

\begin{figure} [H]
	\begin{center}
		\makebox[\textwidth]{\includegraphics[angle=90, height = 0.85\textheight] {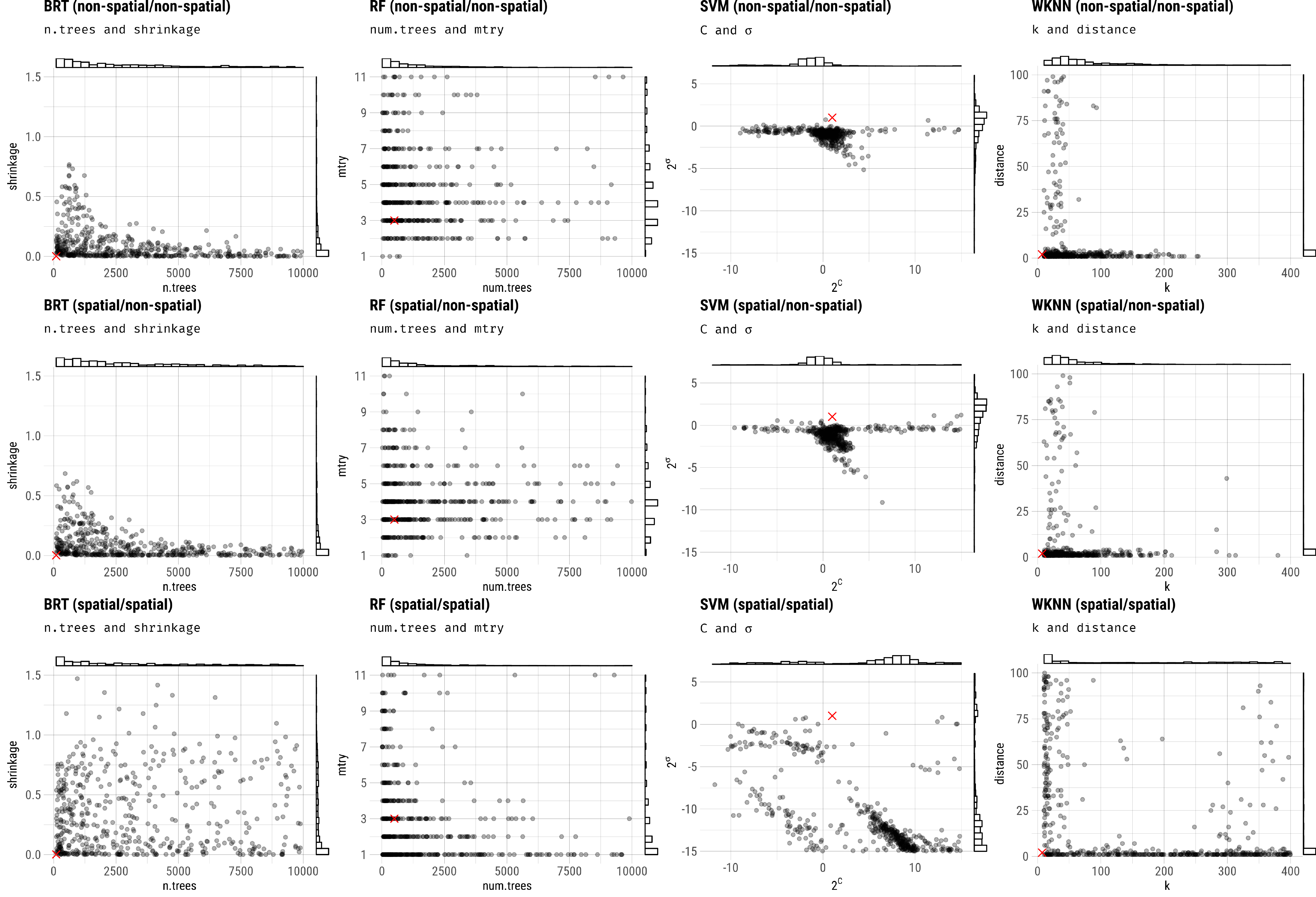}}
		\caption[]{Best hyperparameter settings by fold (500 total) each estimated from 400 random search tuning iterations per fold using five-fold cross-validation. Split by spatial and non-spatial partitioning setup and model type.
			Red crosses indicate default hyperparameter values of the respective model.
			Black dots represent the winning hyperparameter setting out of each random search tuning of the respective fold.}
		\label{fig:best_parameter_combs}
	\end{center}
\end{figure}

There were notable differences in the estimated optimal hyperparameters between the spatial (\emph{spatial/spatial}) and non-spatial (\emph{spatial/non-spatial}, \emph{non-spatial/non-spatial}) tuning settings (\autoref{fig:best_parameter_combs}).
For example when being spatially tuned, the estimated \texttt{$m_{try}$} values of RF mainly ranged between 1 and 3 with \texttt{$m_{try} = 1$} being chosen most often.
In contrast, in a non-spatial tuning situation \texttt{$m_{try}$} was mainly favored between 2 and 4 with \texttt{$m_{try} = 3$} being the mode setting.

\subsection{Predictive performance}
\label{subsec:pred_perf}

\begin{figure} [t!]
	\begin{center}
		\makebox[\textwidth]{\includegraphics[width=\textwidth] {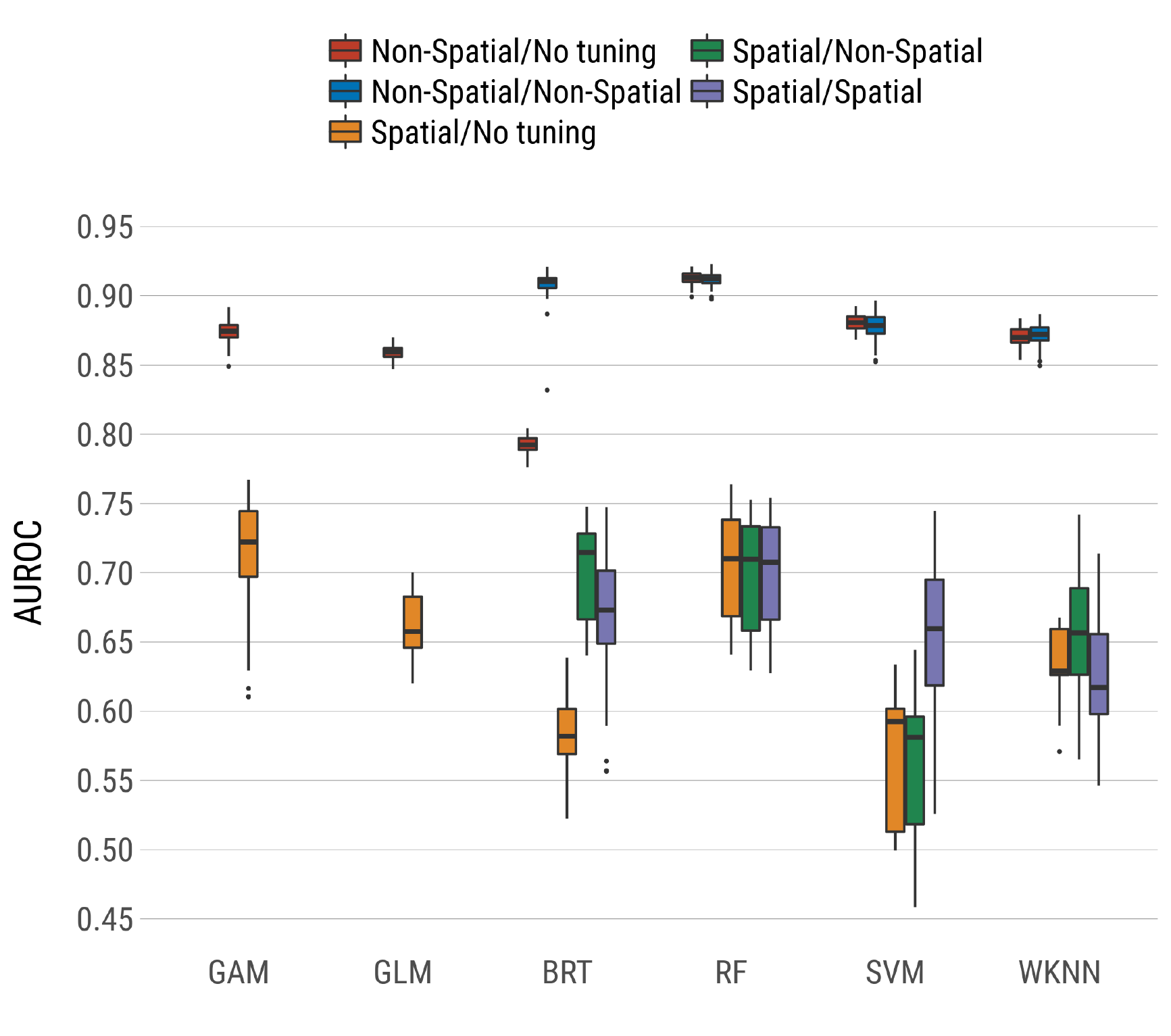}}
		\caption[]{(Nested) \ac{CV} estimates of model performance at the repetition level using 400 random search iterations.
			CV setting refers to performance estimation/hyperparameter tuning of the respective (nested) CV, e.g. "Spatial/Non-Spatial" means that spatial partitioning was used for performance estimation and non-spatial partitioning for hyperparameter tuning.}
		\label{fig:cv_final_boxplots}
	\end{center}
\end{figure}

For the spatial settings (\emph{spatial/spatial} and \emph{spatial/no tuning}), GAM and \ac{RF} show the best predictive performance followed by GLM, SVM and WKNN (\autoref{fig:cv_final_boxplots}).
The absolute difference between the best (RF/GAM) and worst (WKNN) performing model in our setup is 0.081 (mean AUROC, WKNN vs. \ac{RF}/GAM) (\autoref{table:auroc_estimates}).

The tuning of hyperparameters resulted in a clear increase of predictive performance for BRT (0.661 (\emph{spatial/spatial}) vs. 0.587 (\emph{spatial/no tuning}) AUROC) and SVM (0.654 (\emph{spatial/spatial}) vs 0.574 (\emph{spatial/no tuning}) AUROC) (\autoref{table:auroc_estimates}).
The type of partitioning for hyperparameter tuning (spatial (\emph{spatial/spatial}) or non-spatial (\emph{spatial/non-spatial})) only had an substantial impact for SVM (\autoref{fig:cv_final_boxplots}).

Predictive performance estimates based on non-spatial partitioning (\emph{non-spatial/non-spatial} or \emph{non-spatial/no tuning}) are around 24 - 39\% higher, i.e. overoptimistic, compared to their spatial equivalents (\emph{spatial/spatial}).
BRT and WKNN show the highest differences between these two settings (35\% and 39\%, respectively) while the GAM is least affected (24\%).

\begin{table} [t!]
	\centering
	\caption[]{Mean \ac{AUROC} (repetition level) for different 5-fold 100 times repeated cross-validation settings.
		Settings with tuning are based on 400 random search iterations.
		Highest values of each column are highlighted in bold.
		Note that non-spatial performance estimation is over-optimistic.}
	\begin{adjustbox}{max width=\textwidth}
		\begin{tabular}{*{6}{l|c|c|c|c|c|}}
			Performance estimation & \multicolumn{2}{c|}{Non-Spatial} & \multicolumn{3}{c|}{Spatial}                                         \\
			\toprule
			Hyperparameter tuning  & Non-Spatial                      & None                         & Non-Spatial & Spatial    & None       \\
			\midrule
			GLM                    & -                                & 0.859                        & -           & -          & 0.665      \\
			GAM                    & -                                & 0.874                        & -           & -          & \bf{0.708} \\
			BRT                    & 0.908                            & 0.792                        & \bf{0.699}  & 0.671      & 0.583      \\
			RF                     & \bf{0.912}                       & \bf{0.913}                   & 0.698       & \bf{0.699} & 0.704      \\
			SVM                    & 0.878                            & 0.881                        & 0.563       & 0.650      & 0.573      \\
			WKNN                   & 0.872                            & 0.870                        & 0.657       & 0.624      & 0.635      \\
		\end{tabular}
	\end{adjustbox}
	\label{table:auroc_estimates}
\end{table}

\section{Discussion}
\label{sec:discussion}

\subsection{Tuning}

Hyperparameter tuning becomes more and more expensive in terms of computing time with an increasing number of iterations.
Hence, the goal is to use as few tuning iterations as possible to find a nearly optimal hyperparameter setting for a model for a specific data set.
In this respect, random search algorithms are particularly promising in multidimensional hyperparameter spaces with possibly redundant or insensitive hyperparameters (low effective dimensionality; \citep{Bergstra2012}.
These as well as adaptive search algorithms offer computationally efficient solutions to these difficult global optimization problems in which little prior knowledge on optimal subspaces is available.
Bayesian Optimization and F-racing are other approaches that are widely used for optimization of black-box models \citep{Birattari2002, Brochu2010, Malkomes2016}.
In this study, a random search with at least 50 iterations was sufficient for all considered algorithms.

Depending on the data set characteristics, some models (e.g. \ac{RF}) can be insensitive to hyperparameter tuning \citep{Biau2016, Diaz2006}.
As the effect of hyperparameter tuning always depends on the data set characteristics, we recommend to always tune hyperparameters.
If no tuning is conducted, it cannot be ensured that the respective model showed its best possible predictive performance on the data set.

Computing power, especially when conducting a random search, should focus on plausible parameters for each model.
It should be ensured by visual inspection that the majority of the obtained optimal hyperparameter settings does not range closely to the limits of the tuning space.
If the optimal hyperparameter settings are clustered at the edge of the parameter limits, this implies that optimal hyperparameters may actually lie outside the given range.
However, extending the tuning space is not always possible nor practical as numerical problems within the algorithm may occur that may prohibit further extension of the tuning space.
This especially applies to models with a numerical search space (e.g. \ac{SVM}).
In a practical sense one has to question oneself if extending the parameter ranges could possibly result in a significant performance increase and is worth the disadvantage of having an increased runtime.
All these points show the need for a thorough specification of parameter limits for hyperparameter tuning.
As the optimal parameter limits also depend on the dataset characteristics, it is not possible to define an optimal search space for an algorithm upfront.
The chosen parameter limits of this work can serve as a starting point for future analysis but do not claim to be universally applicable.
Users should analyze parameter search spaces of various studies to find suitable limits that match their dataset characteristics.
Within the framework of the \textit{mlr} project a database exists which stores tuning setups of various models from users that can serve as a reference point \citep{mlrhyperopt}.

While in our study no major differences in model performances were found when using spatial versus non-spatial hyperparameter tuning procedures (e.g. 0.03 for \ac{BRT} (0.624 vs. 0.652 AUROC), we recommend using the same (spatial) cross-validation procedure in the inner (tuning) cross-validation step as in the outer (performance estimation).
Generally spoken, hyperparameters from a non-spatial tuning lead to models which are more adapted to the training data than models with hyperparameters estimated from a spatial tuning.
Models fitted with hyperparameters from a non-spatial tuning can then profit from the remaining spatial autocorrelation in the train/test split during performance estimation (compare results of settings \textit{spatial/non-spatial} and \textit{spatial/spatial} of BRT in \autoref{fig:cv_settings_comparison}).
Some software implementations (e.g., the SVM implementation of the \textit{kernlab} package) provide an automated non-spatial CV for hyperparameter tuning.
However, this is only useful for data without spatial and temporal dependencies.

Tuning of RF had no substantial effect on predictive performance in this study.
Nevertheless, the estimated optimal hyperparameters of RF differ for the non-spatial and spatial tuning setting (\autoref{fig:best_parameter_combs}).
In a non-spatial tuning setting, RF will prioritize spatially autocorrelated predictors as these will perform best in the optimization of the \textit{Gini impurity measure} \citep{Biau2016, Gordon1984}.
In this pre-selection \texttt{mtry} values around 3 - 5 are favored because they provide a fair chance of having one of the autocorrelated predictors included in the selection.
At the same time, \texttt{mtry} is low enough to prevent overfitting on the training data which would cause a bad performance on the test set.
This means that mainly the predictors which profit from spatial autocorrelation will be selected.
Although applying these non-spatially optimized hyperparameters on the spatially partitioned performance estimation fold has no advantages in predictive performance compared to using the spatially tuned hyperparameters, the resulting model will have a different structure.
In the spatial tuning setting, mainly \texttt{mtry = 1} is chosen.
This specific setting essentially removes the internal variable selection process by \texttt{mtry} as RF is forced to use the randomly chosen predictor.
Subsequently, on average, each predictor will be chosen equally often and the higher weighting of spatially autocorrelated predictors in the final model (by choosing them more often in the trees) is reduced.
This leads to a more general model that apparently performs better on heterogeneous datasets (e.g. if training and test data are less affected by spatial autocorrelation).

\subsection{Predictive Performance}

In this study we compared the predictive performance of six models using five different \ac{CV} setups (\autoref{subsec:pred_perf}).

Our findings agree with previous studies in that non-spatial performance estimates appear to be substantially "better" than spatial performance estimates.
However, this difference can be attributed to an overoptimistic bias in non-spatial performance estimates in the presence of spatial autocorrelation. (add references)
Spatial cross-validation is therefore recommended for performance estimation in spatial predictive modeling, and similar grouped cross-validation strategies have been proposed elsewhere in environmental as well as medical contexts to reduce bias \citep{Brenning2008, Meyer2018, Pena2015}.

Although hyperparameter tuning certainly increases the predictive performance for some models (e.g. BRT and SVM) in our case, the magnitude always depends on the meaningful/arbitrary defaults of the respective algorithm and the characteristics of the data set.
For SVM, we refrained from using automatic tuning algorithms (e.g. \textit{kernlab} package) or optimized default values (e.g. \cite{e1071}) for all "no tuning" settings.
While the \textit{kernlab} approach clearly violates the "no tuning" criterion, there are no globally accepted default values for $\sigma$ and \texttt{C}.
Subsequently we set both $\sigma$ and \texttt{C} to an arbitrary value of 1.
Naturally, the tuning effect is higher for models without meaningful defaults (such as BRT and SVM) than for models with meaningful defaults such as RF.
Aside from the optimization of predictive performance the aim of hyperparameter tuning is the retrieval of bias-reduced performance estimates.

The biased-reduced outcomes of \ac{RF} (\emph{spatial/spatial} setting) and the GAM (\emph{spatial/no tuning} setting) showed the best predictive performance in our study.
Various other ecological modeling studies confirm the finding that RF is among the best performing models \citep{Bahn2012, Jarnevich2017, Smolinski2016, Vorpahl2012}.
It is noteworthy that the performance of the GLM is close to the one of the GAM and RF for this dataset.

In this work we assume that, on average, the predictive accuracy of parametric models with and without spatial autocorrelation structures is the same.
However, there is little research on this specific topic \citep{Dormann2007b, Mets2017} and a detailed analysis goes beyond the scope of this work.
In our view, a possible analysis would need to estimate the spatial autocorrelation structure of a model for every fold of a cross-validation using a data-driven approach (i.e. automatically estimate the spatial autocorrelation structure from each training set in the respective CV fold) and compare the results to the same model fitted without a spatial autocorrelation structure.
Since we only focused on predictive accuracy in this work, we did not use spatial autocorrelation structures during model fitting for \ac{GLM} and \ac{GAM} to reduce runtime.

Comparing the results of this work to the study of \cite{Iturritxa2014}, an increase in AUROC of \mytilde 0.05 AUROC was observed (comparing the spatial CV result of the GLM from this study to the spatial CV result of \textit{Diplodia sapinea} without predictor \textit{hail} from \cite{Iturritxa2014}).
However, the gain in performance is minimal if predictor \textit{hail\_prob} is removed from the model of this study (0.667 (this work) vs. 0.659 \cite{Iturritxa2014} AUROC).
Subsequently, the influence of the additional predictors \textit{slope}, \textit{soil}, \textit{lithology} and \textit{pH} that were added to this study is negligible small.
The relatively small performance increase of predictor \textit{hail\_prob} (0.667 to 0.694 AUROC) compared to predictor \textit{hail} (0.659 to 0.962 AUROC) from \cite{Iturritxa2014} can be explained by the high correlation of the latter (0.93) with the response.
This inherits from the binary type of the response and predictor \textit{hail}.
The spatially modeled predictor \textit{hail\_prob} of this work is of type numeric (probabilities) and therefore shows a much lower correlation to the response.
In summary, the inclusion of the new predictors increased the predictive accuracy by 0.05 AUROC compared to \cite{Iturritxa2014}.

We want to highlight the importance of spatial partitioning for an bias-reduced estimate of model performance.
If only non-spatial \ac{CV} had been used in this study, the main results of this study would look as follows:
(i) The best model would have been \ac{RF} instead of GAM.
(ii) The predictive performance would have been reported with a mean value of 0.912 AUROC which is \mytilde 0.204 (29\%) AUROC higher than the best bias-reduced performance estimated by spatial \ac{CV} (\emph{spatial/spatial}) (0.708 AUROC, GAM).

\subsection{Other Model Evaluation Criteria}
This work focuses only on the evaluation of models by comparing their predictive performances.
However, in practice other criteria exist that might influence the selection of a algorithm for a specific data set in a scientific field.

Using multiple performance measures suited for binary classification may be a possible enhancement.
However, looking at possible invariances (invariance = not being sensible to changes in the confusion matrix) of performance measures, \cite{Sokolova2009} found that AUROC is among the best suitable measures for binary classification in all tested scenarios.
This is the reason why most model comparison studies with a binary response (e.g. \cite{Goetz2015, Smolinski2016}) only use AUROC as a single error measure.

High predictive performance does not always mean that a model also has a high practical plausibility.
\cite{Steger2016} showed that in the field of landslide modeling, models achieving high AUROC estimates may have a low geomorphic plausibility.

Although the process of automated variable selection is not a criterion that can be compared in a quantitative way, users should always be aware of the selection process of predictor variables when interpreting the plausibility of a model in the ecological modeling field.
While in our case the predictor variables have been selected by expert knowledge, automated variable selection processes (e.g. stepwise variable selection) for parametric models may lead to potentially biased input data \citep{Steger2016}.
As a consequence, the user might receive high performance estimates with unrealistic susceptibility maps \citep{Demoulin2007}.

Another non-quantitative model selection criterion within the spatial modeling field is the surface quality of a predicted map.
Homogeneous prediction surfaces might be favored over predictive power if the difference is acceptable small.
Inhomogeneous surfaces can be an indicator for a poor plausibility of the predicted map, simply caused by the nature of the algorithm (e.g. RF) which splits continuous predictors into classes \citep{Steger2016}.
In comparison a spatial prediction map from a \ac{GAM}, \ac{GLM} or \ac{SVM} shows much smoother prediction surfaces.

\subsection{Model Interpretability}
\label{subsec:mod_interpret}

Although there is an ongoing discussion about the usage of parametric vs. non-parametric models in the field of ecological modeling \citep{Perretti2015}, most studies prefer parametric ones due to the ability to interpret relationships between the predictors and the response \citep{Aertsen2010, Jabot2015}.
However, when interpreting the coefficients of (semi-)parametric spatial models (e.g. \ac{GLM}, \ac{GAM}), spatial autocorrelation structures should be included within the model fitting process (e.g. possible in R with \textit{MASS::glmmPQL()} or \textit{mgcv::gamm()}).
Otherwise, the independence assumption might be violated which in turn might lead to biased coefficients and p-values and hence wrong (ecological) conclusions \citep{Cressie1993, Dormann2007, Telford2005}.

Variable importance information as provided by machine-learning algorithms is only suitable to provide an overview of the most important variables but does not give detailed information about the predictor-response relationships \citep{Hastie2001}.
Using the concept of variable permutation during cross-validation \citep{sperrorest}, \cite{Russ2010a} showed how to analyze variable importance of machine-learning models in the context of spatial prediction.

\section{Conclusion}

A total of six statistical and machine-learning models have been compared in this study focusing on predictive performance.
For our test case, all machine learning models outperformed parametric models in terms of predictive accuracy with \ac{RF} and GAM showing the best results.
The effect of hyperparameter tuning of machine learning models depends on the algorithm and data set.
However, it should always be performed using a suitable amount of iterations and well defined parameter limits.
The accuracy of detecting \textit{Diplodia sapinea} was increased by 0.05 AUROC compared to \cite{Iturritxa2014} with predictor "hail damage at trees" being the main driver.
Spatial \ac{CV} should be favored over non-spatial \ac{CV} when working with spatial data to obtain bias-reduced predictive performance results for both hyperparameter tuning and performance estimation.
Furthermore, we recommend to be clear on the analysis aim before conducting spatial modeling:
If the goal is to understand environmental processes with the help of statistical inference, (semi-)parametric models should be favored even if they do not provide the best predictive accuracy.
On the other hand, if the intention is to make highly accurate spatial predictions, spatially tuned machine-learning models should be considered for the task.
We hope that this work motivates and helps scientists to report more bias-reduced performance estimates in the future.

\section{Acknowledgments}
This work was funded by the EU LIFE project Healthy Forest: LIFE14 ENV/ES/000179.

\section{Appendix}

\appendix
\gdef\thesection{\Alph{section}} 
\makeatletter
\renewcommand\@seccntformat[1]{Appendix \csname the#1\endcsname.\hspace{0.5em}}
\makeatother

\section{Package selection}
\label{app: A}

\subsection{Random Forest}

Several \ac{RF} implementations exist in R.
We used package \textit{ranger} because of its fast runtime.
The \ac{RF} implementation in package \textit{ranger} is up to 25 times faster, taking number of observations as benchmark criteria, and up to 60 times if hyperparameter \texttt{$n_{trees}$} is the benchmark measure, respectively, compared to package \textit{randomForest} \citep{ranger}.
Other packages such as \textit{randomForestSRC}, \textit{bigrf}, \textit{Random Jungle} or \textit{Rborist} lie in between.

\subsection{Support Vector Machine}
Package \textit{kernlab} \citep{kernlab} was chosen in favor of the widely used \textit{e1071} \citep{e1071} package because \textit{kernlab} offers more kernel options.
Other kernels than RBF have been tested partly but not analyzed in detail in this work.

\subsection{Boosted Regression Trees}
For \ac{BRT}, only one implementation exists in R (to our knowledge) in package \textit{gbm} \citep{gbm}.

\subsection{Generalized Linear/Additive Model}
We used the base implementation of \ac{GLM}s in the \textit{stats} package which belongs to the core packages of R.
For \ac{GAM}s, the \textit{mgcv} package was chosen in favor of \textit{gam} because it provides several optimization methods to find the optimal smoothing degree of each variable and the ability to include random effects within the model.
The \textit{mgcv} package lets the user specify different smooth terms and limits for the degree of non-linearity \citep{mgcv}.
By default, the upper limit of parameter $k$, which limits the degree of non-linearity, is set to $k-1$ with $k$ being the number of variables.
Note: It is important to ensure that during optimization $k$ does not hit the upper limit in any of the optimized smooth terms of a predictor variable.
Otherwise, the degree of non-linearity of a predictor variable would be restricted and cannot be modeled accurately.
Subsequently, model performance would not be optimal.
Setting $k$ to a high value relative to the final smoothing degree result leads to highly increased run-time or even convergence problems.

\section{Descriptive summary of numerical and nominal predictor variables}

\begin{table}[H]
\centering
\begingroup\footnotesize
\begin{tabular}{lrrrrrrrrrr}
	\textbf{Variable} & $\mathbf{n}$ & \textbf{Min} & $\mathbf{q_1}$ & $\mathbf{\widetilde{x}}$ & $\mathbf{\bar{x}}$ & $\mathbf{q_3}$ & \textbf{Max} & \textbf{IQR} & \textbf{\#NA} \\
	\hline
	temp              & 926          & 12.6         & 14.6           & 15.2                     & 15.1               & 15.7           & 16.8         & 1.0          & 0             \\
	p\_sum            & 926          & 124.4        & 181.8          & 224.6                    & 234.2              & 252.3          & 496.6        & 70.5         & 0             \\
	r\_sum            & 926          & -0.1         & 0.0            & 0.0                      & 0.0                & 0.0            & 0.1          & 0.1          & 0             \\
	elevation         & 926          & 0.6          & 197.2          & 327.2                    & 338.7              & 455.9          & 885.9        & 258.8        & 0             \\
	slope\_degrees    & 926          & 0.2          & 12.5           & 19.5                     & 19.8               & 27.1           & 55.1         & 14.6         & 0             \\
	hail\_prob        & 926          & 0.0          & 0.2            & 0.6                      & 0.5                & 0.7            & 1.0          & 0.5          & 0             \\
	age               & 926          & 2.0          & 13.0           & 20.0                     & 18.9               & 24.0           & 40.0         & 11.0         & 0             \\
	ph                & 926          & 4.0          & 4.4            & 4.6                      & 4.6                & 4.8            & 6.0          & 0.4          & 0             \\
\end{tabular}
\endgroup
\caption{Summary of numerical predictor variables. Precipitation (p\_sum) in $\mathrm{mm/m^{2}}$, temperature (temp) in \degree C, solar radiation (r\_sum) in $\mathrm{kW/m^{2}}$, tree age (age) in years. Statistics show sample size ($\mathbf{n}$), minimum (\textbf{Min}), 25\% percentile ($\mathbf{q_1}$), median ($\mathbf{\widetilde{x}}$), mean ($\mathbf{\bar{x}}$), 75\% percentile ($\mathbf{q_3}$), maximum (\textbf{Max}), inner-quartile range (\textbf{IQR}) and NA Count (\textbf{\#NA}).}
\label{table:descriptive_summary_numeric}
\end{table}

\begin{table}[H]
\centering
\begingroup\footnotesize
\begin{tabular}{ll|rr}
	\textbf{Variable} & \textbf{Levels}                                                                & $\mathbf{n}$ & $\mathbf{\%}$ \\
	\hline
	diplo01           & 0                                                                              & 703          & 75.9          \\
	                  & 1                                                                              & 223          & 24.1          \\
	\hline
	                  & all                                                                            & 926          & 100.0         \\
	\hline
	\hline
	lithology         & surface deposits                                                               & 32           & 3.5           \\
	                  & clastic sedimentary rock                                                       & 602          & 65.0          \\
	                  & biological sedimentary rock                                                    & 136          & 14.7          \\
	                  & chemical sedimentary rock                                                      & 143          & 15.4          \\
	                  & magmatic rock                                                                  & 13           & 1.4           \\
	\hline
	                  & all                                                                            & 926          & 100.0         \\
	\hline
	\hline
	soil              & \specialcell{soils with little or no profile differentiation                                                  \\ {(Cambisols, Fluvisols)}} & 672 & 72.6 \\
	                  & \specialcell{pronounced accumulation of organic matter in the mineral topsoil                                 \\ {(Chernozems, Kastanozems)}} & 22 & 2.4 \\
	                  & soils with limitations to root growth (Cryosols, Leptosols)                    & 19           & 2.0           \\
	                  & \specialcell{accumulation of moderately soluble salts or non-saline substances                                \\ {(Durisols, Gypsisols)}} & 13 & 1.4 \\
	                  & soils distinguished by Fe/Al chemistry (Ferralsols, Gleysols)                  & 35           & 3.8           \\
	                  & organic soil (Histosols)                                                       & 14           & 1.5           \\
	                  & soils with clay-enriched subsoil (Lixisols, Luvisols)                          & 151          & 16.3          \\
	\hline
	                  & all                                                                            & 926          & 100.0         \\
	\hline
	\hline
	year              & 2009                                                                           & 401          & 43.3          \\
	                  & 2010                                                                           & 261          & 28.2          \\
	                  & 2011                                                                           & 102          & 11.0          \\
	                  & 2012                                                                           & 162          & 17.5          \\
	\hline
	                  & all                                                                            & 926          & 100.0         \\
	\hline
	\hline
\end{tabular}
\endgroup
\caption{Summary of nominal predictor variables}
\label{table:descriptive_summary_non_numeric}
\end{table}

\pagebreak

\section{Additional hyperparameter tuning results}

\begin{figure} [h]
	\begin{center}
		\makebox[\textwidth]{\includegraphics[angle=90, height = 0.6\textheight] {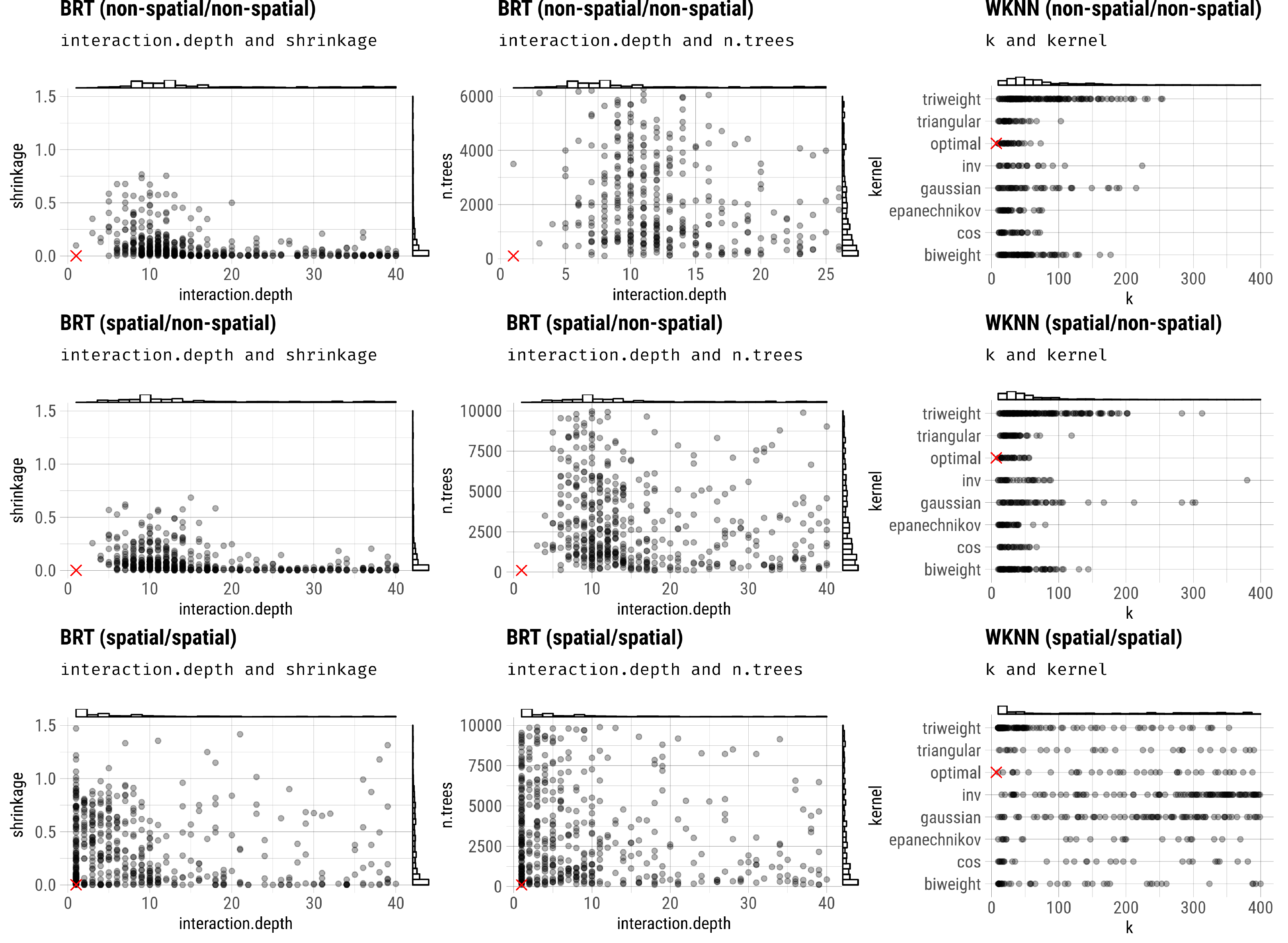}}
		\caption[]{Best hyperparameter settings by fold (500 total) each estimated from 400 random search tuning iterations per fold using five-fold cross-validation.
			Split by spatial and non-spatial partitioning setup and model type.
			Red crosses indicate default hyperparameter values of the respective model.
			Black dots represent the winning hyperparameter setting out of each random search tuning of the respective fold.}
		\label{fig:best_parameter_combs_app}
	\end{center}
\end{figure}

\FloatBarrier

\pagebreak

\section*{References}

\bibliography{../PhD}

\end{document}